\documentclass{article} 

\usepackage{PRIMEarxiv}

\usepackage[utf8]{inputenc} 
\usepackage[T1]{fontenc}    
\usepackage{hyperref}       
\usepackage{url}            
\usepackage{booktabs}       
\usepackage{amsfonts}       
\usepackage{nicefrac}       
\usepackage{microtype}      
\usepackage{lipsum}
\usepackage{fancyhdr}       
\usepackage{graphicx}       
\graphicspath{{media/}}     

\usepackage[numbers]{natbib}
\usepackage[table]{xcolor}
\usepackage{multirow}

\usepackage[acronym]{glossaries}
\newacronym{AMODEL}{NBCSAuthor}{NBC-Softmax :  Darkweb Author fingerprinting and migration tracking}  
\newacronym{ALOSS}{NBC-Softmax}{Negative Block Contrastive Softmax} 
\title{\acrlong{AMODEL}}

\author{
    Gayan K. Kulatilleke, Shekhar S. Chandra \& Marius Portmann \\
    University of Queensland, Brisbane, Australia \\
    \texttt{\{g.kulatilleke@uqconnect, shekhar.chandra@uq, marius@itee.uq\}.edu.au}
}

\begin{document}
\maketitle
\begin{abstract}
Metric learning aims to learn distances from the data, which enhances the performance of similarity-based algorithms.
An author style detection task is a metric learning problem, 
where learning style features with small intra-class variations and larger inter-class differences is of great importance to achieve better performance.
Recently, metric learning based on softmax loss has been used successfully for style detection.
While softmax loss can produce separable representations, its discriminative power is relatively poor.
In this work, we propose \acrshort{ALOSS}, a contrastive loss based clustering technique for softmax loss, which is more intuitive and able to achieve superior performance. Our technique meets the criterion for larger number of samples, thus achieving block contrastiveness, which is proven to outperform pair-wise losses. It uses mini-batch sampling effectively and is scalable.
Experiments on 4 darkweb social forums, with \acrshort{AMODEL} that uses the proposed \acrshort{ALOSS} for author and sybil detection, shows that our negative block contrastive approach constantly outperforms state-of-the-art methods using the same network architecture.

Our code is publicly available at : https://github.com/gayanku/NBC-Softmax
\end{abstract}

\keywords{block contrastive \and metric learning \and darkweb \and classification \and fingerprinting \and sybil \and multitask \and user analysis \and authorship}


\section{Introduction}
The darkweb is a subset of the deep web that is intentionally hidden, requiring specific privacy focused software and communication protocols, such as the Tor router \footnote{https://www.torproject.org/} or Invisible Internet Project (I2P) \footnote{https://geti2p.net/} for access \cite{kumar2020edarkfind}. This infrastructure guarantees anonymity and protection from surveillance and tracking \cite{manolache2022veridark,munksgaard2016mixing}.
Darkweb includes illicit drug trade, money laundering, counterfeit goods, information for sale and many other illicit services \cite{biryukov2014content,munksgaard2016mixing}. Predominantly, pseudo-anonymous crypto currency is the preferred means of payment \cite{biryukov2014content}.
Despite wide law enforcement crackdowns, darkweb market users are resilient to closures \cite{munksgaard2016mixing}. Rapid migrations to newer markets occur when one market shuts down \cite{elbahrawy2020collective}, and the total number of users remains relatively unaffected. Non-English and multilingual darkweb markets have been increasing in number since 2013 \cite{ebrahimi2018detecting}.

The majority of the darkweb markets have an associated forum where users can interact, discuss vendors and products, provide feedback, and build trust \cite{lorenzo2018know}. Interactions on these forums are facilitated by means of text posted by their users. The only link between a real individual’s identity and activities in the darkweb is the username linked to a post \cite{maneriker2021sysml}. 
This makes the analysis of these forums and post content a compelling problem \cite{maneriker2021sysml}. However, these forums discussions remain underutilized for identity analysis \cite{munksgaard2016mixing}. Furthermore, most often, malicious users maintain multiple accounts (or Sybil accounts) within and across different crypto markets \cite{kumar2020edarkfind,tai2019adversarial}.

\textbf{User fingerprinting}
Existing research indicates the average lifespan of darkweb markets to be around 8 months \cite{acnikevivcs2021relationship}. As a result, in order to maintain reputation and be able to quickly transition to a new darkweb market and ensure success, vendors tend to maintain their reputation between markets by using their username as a brand \cite{acnikevivcs2021relationship}. 

Two accounts of the same vendor are most likely to have similar usernames on different darkweb markets or forums, while two accounts of same vendor in the same marketplace tend to be different \cite{acnikevivcs2021relationship}. At present, (2022), relocation to other darkweb markets and forums is the key issue for law enforcement when dealing with darkweb cybercrime \footnote{Europol, 2022}. Thus, more efforts are needed to develop insight generation and forensic tools to help analysts \cite{manolache2022veridark}. The most significant digital footprint in the darkweb, for forensic analysis, is a user’s post history \cite{manolache2022veridark}. Linking different accounts operated by the same individual facilitates building a more accurate and refined profile of activities, track users across forums and properly assess behavior of darkweb users \cite{kumar2020edarkfind}. Linking different accounts aid accurately evaluate substance volumes advertised across the different crypto markets by each vendor \cite{kumar2020edarkfind}.

Authorship profiling and authorship attribution are important areas in forensic linguistics~\cite{manolache2022veridark}. However, traditional techniques rely upon long, relatively clean and jargon free text to train language models. These techniques cannot be effectively applied for short, nosy text corpora, in a heavily anonymized environment \cite{maneriker2021sysml}.
Recent work has been successful in using deep neural networks specifically design to address the aforementioned scarcity of easily identifiable `signature' features; \cite{shrestha2017convolutional} used character and word-level modelling on shorter text, \cite{andrews2019learning} combined subreddits (considering them as context) with posts to address the shortness of Reddit posts and most recently, \cite{maneriker2021sysml} used a softmax based supervised metric loss and multitask learning to achieve state-of-the-art.

The author style detection task can be formulated as a metric learning problem, where learning style features with small intra-class variations and larger inter-class differences is of great importance to achieve better performance \cite{wang2018additive}.
Current state-of-the-art for (darkweb) social forum author identification, SYSML~\cite{maneriker2021sysml} as well as many other classification-based detection and recognition tasks are mostly based on the widely used softmax loss \cite{wang2018additive}.  While softmax loss optimizes the inter-class differences by separating different classes well, it is not effective at reducing the intra-class variation to make features of the same class compact in representation space~\cite{wang2018additive}. While softmax loss can produce separable representations, its discriminative power is relatively poor \cite{wen2016discriminative}.
In author identification tasks, representations needs to  be not only separable but also discriminative. Features should be discriminative and generalized sufficiently in order to identify new unseen authors, as well as be able to show two authors, differently labelled, can in fact be the same sybil.
As a result, many new loss functions have been proposed to minimize the intra and inter class variation \cite{wang2018cosface,deng2019arcface,wang2019multi}.

Most of these variants are based on maximizing the separation margin of the class vectors and rely on large margin classification. 
However, in user detection, where the ground truth can have a higher degree of human annotation errors (especially in case of sybils), compacting the intra-class variation may impact the optimization. Further, above mentioned methods are only applicable to supervised deep metric learning. In an unsupervised setting, contrastive learning approaches achieve superior performance across multiple domains \cite{kulatilleke2022efficient,kulatilleke2022scgc}. Discriminative approaches based on contrastive learning has been outstandingly successful in practice \cite{wang2020understanding}, achieving superior results or at times even outperforming its supervised learning counterparts \cite{logeswaran2018efficient}.

Hence, in this work we take the opposite approach. Rather than attempt to compact intra class variation or force strict larger margins as in prior work, we distance the weight vector of each class based on similarity using proxy (or prototype, or surrogate) block contrastive learning. Also, different to typical approaches that contrast negative samples with positives, we only use the negative samples, \textit{and} even then, only the negative sample aggregates. Further, we contrast at a lower dimensional latent space which forces the network to extract common factors of variation from the data.

Note that the application of contrastive learning to supervised metric loss is non-trivial. 
Both contrastive loss (contrasting pairs of samples) and triplet loss (contrasting a triplet that where the anchor is contrasted with a positive and a negative) based metric loss suffer from dramatic data expansion due to expensive mining strategies used to obtain pairs or triplets for training. Additionally, the performance is extremely sensitive to hyperparameters and learning strategies \cite{wang2018additive}. 

Our aforementioned approach does not require sampling, is robust to hyperparameter variations and is scalable. It demonstrates increasing performance with larger batch size and length. We are motivated by the effectiveness of contrastive learning, the availability of label information, and the recent work in surrogate proxies \cite{kulatilleke2022scgc,kulatilleke2022efficient}. While \cite{gunel2020supervised}, using cross-entropy with supervised contrastive loss is the most similar to our work, it is not block contrastive, and is based on traditional negatives and positive pairs, where as our \acrshort{ALOSS} only requires negative clusters.

To summarize, our main contributions are:
\begin{itemize}
\item We introduce a novel loss, \acrfull{ALOSS} to improve inter-cluster discrimination of traditional softmax loss.
\item we provide a thorough supervised metric learning problem formulation relating this softmax based learning.
\item we provide a detailed study of the hyperparameters of our \acrshort{ALOSS} including comparison with alternative metric learning approaches such as CosFace (CF)~\cite{wang2018cosface}, ArcFace (AF)~\cite{deng2019arcface}, and MultiSimilarity (MS)~\cite{wang2019multi}.
\item Experiments on 4 darkweb social forums for author and sybil detection shows that our \acrshort{ALOSS} constantly outperforms state-of-the-art methods using the same network architecture. 
\item Importantly, we show that our \acrshort{ALOSS} can significantly improve detection by combining multiple darkweb datasets from different sources and effectively link sybil accounts achieving up to 49\% over prior work.
\end{itemize}

Our work relates to darkweb forensic investigations and author profiling with the intention of detecting sybils, or users with multiple identities. The learnt representations can also be used for possible author identification (i.e., who authored the content?, a selection from a closed set task), verification (did the author write a given content, a binary classification) and other downstream tasks.

\section{Preliminaries and Related works}
In this section, we review two different families of deep metric learning approaches, namely classification and block-wise contrastive losses. Specifically, from an authorship attribution prospective, in an embedding network that maps inputs to a high dimensional space, classification loss is a function of the embedding (authored content) and its category label (author). Traditional (pairwise) contrastive loss, on the other hand, is a function of the similarity or distance between two embeddings (authored content) of the corresponding pairwise (author) labels. 

\subsection{Traditional Softmax classification loss}
Softmax and most of its variants such as CosFace~\cite{wang2018cosface} and ArcFace~\cite{deng2019arcface}, attempt to increase the margin, typically by modifying $\theta$ and reply on large margin classification. This can also be effectively applied on other domains \cite{munksgaard2016mixing,patel2022recall}. 

CenterLoss \cite{wen2016discriminative} minimizes the intra-class distances by computing the centres by averaging the features of the corresponding classes and penalizing samples that are further away in embedding space, essentially adding a regularization term to softmax. As in Centerloss, class centers closer to the center benefit less from regularization, Sphereface \cite{liu2017sphereface} modified softmax to normalize the penalty by enforcing the class centers to be on the surface of a unit hypersphere, equidistant from the center.  However, The decision boundary of the SphereFace, defined over the angular space by  $\cos(\mu \theta_1) = \cos(\theta_2)$  is hard to optimize due  the  cosine function’s nonmonotonicity, which CosFace \cite{wang2018cosface}.  solved by adding a decision margin $cos(\theta)+m$. ArcFace \cite{deng2019arcface} addresses the same sphereface limitations. However instead of defining the margin in the cosine space, ArcFace uses $cos(\theta+m)$, defining the margin in angle space directly. Figure~\ref{fig_NCEBoundries} shows decision boundaries of these different loss functions.

\subsection{Traditional (non-block) Contrastive loss}
Many embedding models \cite{wang2020understanding, kulatilleke2022scgc} use contrastive objectives, which encourage representation of positive pairs (distinguish similar representations) to be similar, and make representations of negative pairs further apart in embedding space \cite{wang2020understanding}. 
In unsupervised settings, with no labels available, augmented samples are used as positives, and negatives are randomly drawn openly \cite{chen2020simple}. Maximizing agreement between such two augmented samples by assuming IID data is often used in vision, a method pioneered by SimCLR's \cite{chen2020simple} NCE approach. These discriminative approaches based on contrastive learning have been outstandingly successful in practice \cite{wang2020understanding}, reporting state-of-the-art results \cite{chen2020simple} as well as outperforming supervised learning \cite{logeswaran2018efficient} in multiple domains. In the text domain, \cite{logeswaran2018efficient} proposed QT, a SOTA contrastive language model which works in sentence-space rather than word-space, which is an order of magnitude faster than previous methods, achieving better performance simultaneously. \cite{kulatilleke2022scgc} proposed SCGC, a lean contrastive model based on MLP that could handle text, graph and image data. 

Recently, supervised contrastive learning \cite{khosla2020supervised} has been able to extend and improve on unsupervised contrastive learning by exploiting label information in order to choose positive and negative samples. It outperforms cross entropy loss achieving state-of-the-art results
on ImageNet. 
\cite{gunel2020supervised} proposed a fine-tuning objective combining softmax with a supervised contrastive learning term to significantly improve the RoBERTa-Large (text) baseline, by pulling examples from the same class closer and pushing different classes further apart.

\subsection{Block Contrastive loss}
Block-wise contrastive losses, using aggregated sample (or blocks) instead of a collection of pairs has a strictly better bound due to Jensen’s inequality getting tighter, and is superior compared to their equivalent of element-wise contrastive counterparts \cite{arora2019theoretical}. \cite{kulatilleke2022scgc} proposed PAMC, a block contrastive approximation with efficient surrogate negative computation, that works on multiple data modalities including graphs,text and images and avoids the need for expensive sample mining.

\subsection{Authorship attribution}
Authorship attribution identifies an unknown text’s author from a set of candidates~\cite{kumar2020edarkfind}, useful from plagiarism detection to forensic linguistics. Convolutional neural networks (CNNs) have been used for text classification \cite{kim2014convolutional}. \cite{shrestha2017convolutional} showed that for short text sequences of character n-grams, CNNs outperformed other models for author attribution. Byte-level tokenization enables having vocabularies across data in multiple languages~\cite{maneriker2021sysml} and has been used cross multilingual social media data~\cite{andrews2019learning}. 
Using chronologically sequences of actions (post content and other meta data), termed episodes, ~\cite{andrews2019learning} created user invariant features using a discardable surrogate auxiliary metric learning objective based on angular margin~\cite{deng2019arcface}. eDarkFind~\cite{kumar2020edarkfind} uses multiple sources with BERT, a knowledge graph of drug substances and location representation, among others, fused, to obtain representations. We use the darkweb datasets from \cite{munksgaard2016mixing}, which has been used by \cite{acnikevivcs2021relationship,maneriker2021sysml}.

\section{Approach}
The metric learning task can be abstracted to a model $f(\theta)$ and a suitable loss $L(\theta)$ used for its optimization. To better understand our \acrshort{ALOSS}, $L(\theta)$, we first define the supervised learning problem in a local context, review of the original softmax loss with our intuition and motivation, followed by the mathematical definitions. Thereafter, we describe the darkweb author identification model, $f(\theta)$ based on SYSML \cite{maneriker2021sysml} on which we benchmark \acrshort{ALOSS}. In this paper, better performance refers to better $MRR$ and better $R@k, \forall k \in \{1,2,5,10\}$. 

\subsection{Problem Setting of Supervised Metric Learning}
In this section, we review the distance metric learning problem.

Given a non-empty set $X$, \emph{distance} or \emph{metric} over $X$ is a map $d \colon X \times X \to \mathbb{R}$ satisfying the following properties:
\begin{enumerate}
\item Coincidence: $d(x,y) = 0 \iff x = y \quad \forall \quad x, y \in X$.
\item Non negativity: $d(x,y) \ge 0 \quad \forall \quad  x,y \in X$.
\item Symmetry: $d(x,y) = d(y,x) \quad \forall \quad x,y \in X$.
\item Triangle inequality: $d(x,z) \le d(x,y) + d(y,z) \quad \forall \quad x,y,z \in X$.
\end{enumerate}
The ordered pair $(X,d)$ is called a \emph{metric space}.

In the most generic use case of metric learning, a dataset $\mathcal{X} = \{x_1,\dots,x_N\}$ is available, on which certain similarity measures between different pairs or triplets of data are collected. These similarities are determined by the sets
\begin{align*}
  S &= \{(x_i,x_j) \in \mathcal{X}\times\mathcal{X} \colon x_i \text{ and } x_j \text{ are similar.} \}, \\
  D &= \{(x_i,x_j) \in \mathcal{X}\times\mathcal{X} \colon x_i \text{ and } x_j \text{ are not similar.} \}, \\
  R &= \{(x_i,x_j,x_l) \in \mathcal{X}\times\mathcal{X}\times\mathcal{X} \colon x_i \text{ is more similar to } x_j \text{ than to } x_l. \}.
\end{align*}
Solving the optimization problem
\[ \min_{d \in \mathcal{D}} \ell(d,S,D,R) .\]
with some metric loss function $\ell$, provides the distances (represented by embeddings on some vector space) that best adapt to the criteria specified by the similarity constraints.

In supervised metric learning, additionally, there is a corresponding list of discrete and finite set of labels $y_1,\dots,y_N$ for every $\mathcal{X}$. 
By considering the sets $S$ and $D$ as sets of pairs of same-class (positive) samples and different-class (negative) samples respectively, supervised learning can be incorporated into distance metric learning.  There are two main methods for establishing these sets, \textit{global} and \textit{local} \cite{suarez2018tutorial}.
Following \cite{suarez2018tutorial}, In the global approach, 
\begin{align}
  S &= \{(x_i,x_j) \in \mathcal{X}\times\mathcal{X} \colon y_i = y_j\}, \\
  D &= \{(x_i,x_j) \in \mathcal{X}\times\mathcal{X} \colon y_i \ne y_j\}. 
\end{align}
Alternatively, in the local approach:
\begin{equation}
    S = \{(x_i, x_j) \in \mathcal{X}\times\mathcal{X} \colon y_i = y_j \text{ and } x_j \in \mathcal{U}(x_i)\},
    \label{EQU_Local}
\end{equation}
incorporates a \emph{neighborhood} of $x_i$, $\mathcal{U}(x_i)$.

Essentially, the goal is to train a deep neural network $f_{\theta}(\cdot)\, \colon \mathcal{X} \to \mathbb{R}^n$  and a distance $ \mathcal{D}\, \colon \mathbb{R}^n \to \mathbb{R}$ such that for samples $ x_1, x_2 \in \mathcal{X}$ and corresponding labels $ y_1, y_2 \in \mathcal{Y}$, $\mathcal{D}\left(f_{\theta}(x_1), f_{\theta}(x_2)\right)$ produces smaller values if the labels $ y_1, y_2 \in \mathcal{Y}$ are equal, and larger values otherwise. A deep metric learning problem consists of determining the architecture for $f_{\theta}$ and appropriate function $ \mathcal{L}(\theta)$ for optimization.

\subsection{Drawbacks and limitations of softmax loss}
Note that the global approach, described above in Equation~\ref{EQU_Local}, requires  considering the entire training set in each iteration, which is inefficient, computationally prohibitive and not scalable. While this is not required in the local approach, there is the need for incorporating the neighborhood $\mathcal{U}(x_i)$. Softmax loss works at sample level and cannot adequately incorporate $\mathcal{U}(x_i)$ in the loss computation.

Furthermore, in detection and recognition tasks, classes are typically within the training set, i.e., it is a close-set identification, and predicted labels dominate the performance. As a result, softmax loss to be able to optimize the network for classification problems, leading to separable representations \cite{wen2016discriminative}. However, for author identification tasks, representations need to be not only separable but also discriminative. Features should be discriminative and generalized sufficiently in order to identify new unseen authors, as well as be able to show two authors, differently labelled, can in fact be the same sybil. Discriminative power distinguishes features by compact intra-class variations and separable inter-class differences \cite{wen2016discriminative}

\subsection{In to softmax classification loss}
Essentially, classification learns a decision boundary to separate classes. 
The original softmax loss is given by:
\begin{equation}
\begin{aligned}
\mathcal{L}_{SCE} & = -\frac{1}{n}\sum_{i=1}^n{\log\frac{e^{W_{y_i}^T {f}_i}}{\sum_{j=1}^{c}{e^{W_j^T {f}_i}}}} \\
& = -\frac{1}{n}\sum_{i=1}^n{log\frac{e^{\|W_{y_i}\| \|{f}_i\|cos(\theta _{y_i})}}{\sum_{j=1}^{c}{e^{\|W_j\| \|{f}_i\| cos(\theta_{j})}}}},
\label{EQ_SOFTMAX}
\end{aligned}
\end{equation}
where $f_i$ is the $i$-th sample input and $W_j$ is the $j$-th column of the final fully connected layer respectively and $\theta$ is the angle between the weight $W$ and feature $x$. $W_{y_i}^T {f}_i$ is also known as the $i$-th sample's target logit \cite{wang2018additive}. As shown in Figure~\ref{fig_THEORY}, for a 2 class classification, the weight vectors are $W_1$ and $W_2$. Note that at the decision boundary, $|W_1| = |W_2|$. For a sample $x$ from class 1, for correct classification
\begin{equation}
    W_1\cdot x >  W_2 \cdot x 
    \label{EQ_A}
\end{equation}
or alternatively,
\begin{equation}
    \|W_{1}\| \|{f}_i\|cos(\theta _{1}) > \|W_{1}\| \|{f}_i\|cos(\theta _{2})
    \label{EQ_B}
\end{equation}
Softmax and most of its variants such as CosFace~\cite{wang2018cosface} and ArcFace~\cite{deng2019arcface}, attempt to increase the margin, typically by modifying $\theta$ and rely on large margin classification. 

Differently, we replace $W_1$ with $W_1^*$ and $W_2$ with $W_2^*$, given class1 $\ne$ class2, using negative contrastive loss to force the weights away from each other, thus indirectly, \textit{softly} increasing $\theta$ based on a learnt similarity. In this work we use cosine similarity.  

Classification losses, in contrast to pairwise or block losses, typically optimize embeddings independently \cite{patel2022recall}.

\subsection{\acrshort{ALOSS} : Supplementing softmax with negative contrastive inter-class clusters}
Traditional softmax loss is defined as:
\begin{equation}
l_{\operatorname{SCE}}=- { \frac{1}{b} \sum_{i=1}^{b} \log \frac{ \exp( \theta(z_{i,y_i}) )}{  \sum_{j=1}^{C} \exp( \theta(z_{i,y_j})) } },
\label{EQ_SEC}
\end{equation}
where $b$ is the batch size, $C$ represents the classes and $\theta$ is a deep neural network.

We define the negative block contrasting term as:
\begin{equation}
l_{\operatorname{NEG}} = \frac{1}{ | C\sp{\prime} |} \log{
\sum_{a=1}^{C\sp{\prime}} \sum_{b=1}^{C\sp{\prime}} \mathbf{1}_{[a \neq b]} 
\exp \left(\operatorname{sim}\left(\boldsymbol{\hat{\mu}}_{a}, \boldsymbol{\hat{\mu}}_{b}\right) \cdot \tau\right)
}\label{EQ_NEG}
\end{equation}
where $\operatorname{sim()}$ is some similarity function. In this work, we use cosine similarity. $\tau$ is the customary temperature hyperparameter \cite{khosla2020supervised,kulatilleke2022scgc,kulatilleke2022efficient} found in contrastive loss. We only use non-empty labels, denoted by $C\sp{\prime}$. 

Note that, $l_{\operatorname{NEG}}$ is free from $i$ terms. Computing $|C\sp{\prime}| \times |C\sp{\prime}|$ is more efficient than $|b| \times |C|$. Further, it is also possible to compute $l_{\operatorname{NEG}}$ once per epoch for all batches. We define the terms being used, $\hat{\mu_c}$ and $C\sp{\prime}$ in Equation~\ref{EQ_MUHAT} and Equation~\ref{EQ_CHAT} respectively. 
\begin{equation}
    \hat{\mu_c}=\frac{1}{| \sum_{i=1}^{b} \mathbf{1}_{[ y_i = c]} |}
    \sum_{i=1}^{b} \mathbf{1}_{[ y_i = c]} z_i
    \label{EQ_MUHAT}
\end{equation}

\begin{equation}
    C\sp{\prime} = \{ \hat{\mu_k} |  k \in \{1, \dots, C\} \land | \sum_{i=1}^{b} \mathbf{1}_{[ y_i = k]}| > 0 \}
    \label{EQ_CHAT}
\end{equation}

Thus, with $\alpha$ as the hyperparameter related to balancing the effects (which in this work we keep fixed at 0.5), the final \acrshort{ALOSS} is:
\begin{equation}
    l_{\operatorname{combined}} =  \alpha l_{\operatorname{SCE}} + (1 - \alpha) l_{\operatorname{NEG}}
    \label{EQ_LCOMBINED}
\end{equation}

\subsection{Geometric interpretation}
\begin{figure}
    \centering \includegraphics[width=0.7\columnwidth]{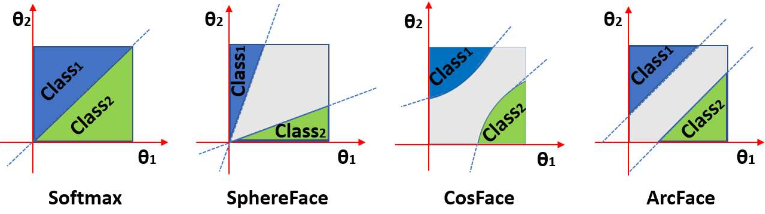}
    \caption{A visual overview of different softmax based metric learning approaches showing decision boundaries. (Image source \cite{deng2019arcface})}
    \label{fig_NCEBoundries}
\end{figure}
\begin{figure}
    \centering \includegraphics[width=0.5\columnwidth]{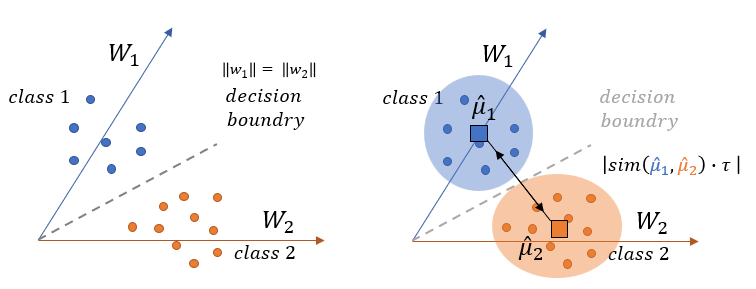}
    \caption{Comparison between the traditional softmax loss with our \acrshort{ALOSS}. We use similarity of different classes, represented and managed by $\hat{\mu}$ to force apart the weight vectors $W$, instead of imposing any soft or hard margins.}
    \label{fig_THEORY}
\end{figure}
Figure~\ref{fig_NCEBoundries} provides a geometrical interpretation of popular softmax loss variants, with different colors representing feature space from different classes. It can be seen that features learned by traditional softmax is separable, but less discriminative and classes cannot be  classified simply via angles, while in the modified variants, margins are used for separation as well as compacting the intra-class samples. However, all these methods assume a 'fixed' set of weight vectors, in this case indicated by $\theta_1, \theta_2$

\subsubsection{Alignment and uniformity}
Alignment and uniformity are key properties of contrastive loss, where alignment encourages an encoder to assign similar features to similar samples, and uniformity enforces the feature distribution to preserve maximal information \cite{wang2020understanding}. Uniformly distributing points on a hyperspace is defined as minimizing the total pairwise potential w.r.t. a certain kernel function and is well-studied \cite{wang2020understanding}.

Our approach focuses on enhancing the uniformity. \acrshort{ALOSS} achieves this objective by using $s i m()$ as the kernel function.
As shown in Figure~\ref{fig_THEORY}, \acrshort{ALOSS} enforces a loss penalty on the weights $W_1, W_2$ or the 2 classes based on some latent notion of similarity, guided with labels. 

\subsection{Author detection framework}
\begin{figure}
    \centering \includegraphics[width=0.8\columnwidth]{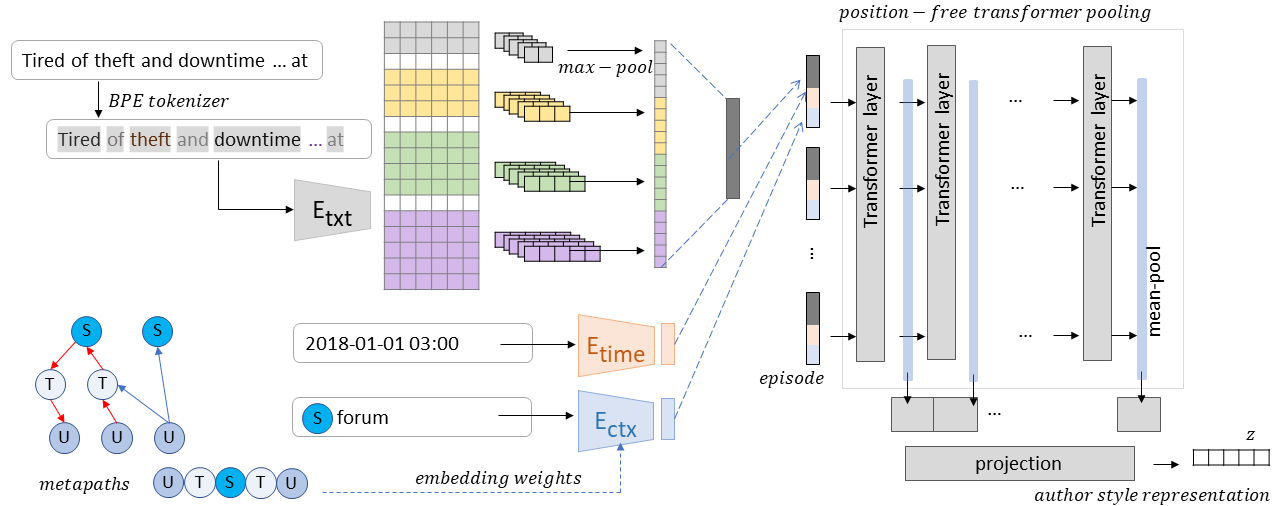}
    \caption{Episodes composes of sets of text (forum post), time (of the post) and forum context (taken from the meta-path based heterogeneous graph graph) triplets. Each episode becomes an input to the transformer layer. We use a set size of 5 episodes to counter the short text nature of forum posts and also to be able to capture author style information. The final representation, $z$ is used the used for downstream tasks. The model is based on SYSML\cite{maneriker2021sysml}. The text extraction model is based on \cite{kim2014convolutional}, the context module uses metapath2vec from \cite{dong2017metapath2vec}}
    \label{fig_SYSMLNetwork}
\end{figure}

In this section we provide an overview of the author detection model, i.e. $f(\theta)$, and details of the forum content it uses as features for style representation learning. Note that this is heavily inspired by SYSML \cite{maneriker2021sysml} for multitask and context learning, uses social media user representations learning by combination of multiple posts by each user proposed by \cite{andrews2019learning} for a singletask, and it adopts the proxied block contrastive loss PAMC \cite{kulatilleke2022efficient}. 
Following \cite{maneriker2021sysml,andrews2019learning}, we combine multiple posts from a user. Following \cite{maneriker2021sysml}, we supplement user posts with time (day of week) and sub-forum information and call this triplet, (i.e., post text, time, sub-forum context) an episode. A set of episodes, with an episode\_length of $L$ is passed to a deep network $f_{\theta}$ which provides a combined representation $z \in \mathbb{R}^E$. 
The model can be trained on a metric learning tasks $g_{\phi}: \mathbb{R}^E \xrightarrow[]{} \mathbb{R}$; \cite{maneriker2021sysml} used softmax loss, we use \acrshort{ALOSS}. The metric learning task ensures episode sets from the same author are placed closer in some latent, semantic embedding space. 
Figure~\ref{fig_SYSMLNetwork} shows the main components of the deep neural architecture, which we will very briefly describe in the following sections for completeness. For full details we direct the reader to \cite{maneriker2021sysml}. 

\subsubsection{Post text}
User post is tokenized using byte-level tokenizer (BPE), one-hot encoded by an embedding matrix $E_{txt}$ of dimensions $|V| \times d_t$ with a $V$ token vocabulary and $d_t$ embedding dimension to obtain a $T_0$, $T_1$, $\dots, T_{n-1}$ token sequence with an dimension $n \times d_t$. 
According to \cite{maneriker2021sysml}, as most darkweb markets are multi-lingual (e.g., BMR contained a large German community), using BPE allows a shared vocabulary to be used across multiple datasets with relatively less out-of-vocab tokens. 
Following \cite{kim2014convolutional}, $f_n$ sets of sliding window filters with $F = \{2, 3, 4, 5\}$ sizes are used to obtain feature-maps, which are max-pooled to obtain a  $|F| \times f_n$ dimensional embedding, passed through a dropout layer to avoid overfitting  and  finally projected to $d_{txt}$ dimensions using a fully connected layer. 

\subsubsection {Time (day of week)}
Different darkweb markets in the dataset are of different time granularities. Additionally, there is a possibility of the server time being different or in an unknown time zone.  We follow \cite{maneriker2021sysml} and use only the day of the week $0,\dots,7$ from each post, via a time embedding matrix $E_{time}$ with a dimension of $d_{time}$.

\subsubsection{Sub-forum context}
Darkweb forums have sub forums that focus on specific topics or areas of interest, which can be used to obtain a post's associated context. \cite{andrews2019learning} used randomly initialized and optimized embeddings from one-hot encoded subreddits for context of Reddit posts.  \cite{maneriker2021sysml} extended this by initializing the forum embedding matrix, $E_{ctx}$ with a meta path based heterogeneous graph optimization, which we adopt. Specifically, following metapath2vec~\cite{dong2017metapath2vec}, a graph with four node types, user (U), subforum (S), thread (T), and post (P) are used. Edges indicate different relationships, for example U-T is a post to a new thread, U-P is a reply to existing post. We use the 7 meta paths defined by \cite{maneriker2021sysml}, namely UPTSTPU, UTSTPU, UPTSTU, UTSTU, UPTPU, UPTU, and UTPU to capture behavioral semantics.  Metapath2vec uses meta-path based random walks and aims to:
\begin{align*}
            \arg\max\limits_\theta \prod\limits_{v\in V} \prod\limits_{t\in T_v} \prod\limits_{c_t \in N_t(v)} p(c_t|v; \theta)
\end{align*}  
For a heterogeneous graph $G = (V, E, T)$, a node $v$ and edge $e$ contain an associated edge of  `relationship type' $T_i \in T$, with a mapping $\phi(v): V \rightarrow T_V$, $\psi(e): E \rightarrow T_E$, satisfying $|T_V| + |T_E| > 2$. Essentially, Metapath2vec maximizes the probability taking into consideration the multiple types of nodes and edges using the heterogeneous skip-gram model to perform node embeddings ~\cite{dong2017metapath2vec}. We obtain the embeddings for the sub-forums to initialize weights in embedding metric $E_{ctx}$ and obtain time embeddings of $d_{time}$.  

\subsubsection{Episodes}
Episodes of dimension $d_e = d_{txt}+ d_{time} + d_{ctx}$ are obtained by concatenating post, time and context. We use a position-free transformer~\cite{devlin-etal-2019-bert,vaswani2017attention,kulatilleke2021fdgatii} where episodes are elements $1,\dots,L$ of the input sequence.
The transformer uses self-attention to capture the dependency between episodes and is also capable of parallel computation with high efficiency.
We obtain $z = f_{\theta}(\{ (\operatorname{txt, time, context})\}^L)$.

\subsection{Learning}
\begin{figure}
    \centering \includegraphics[width=0.8\columnwidth]{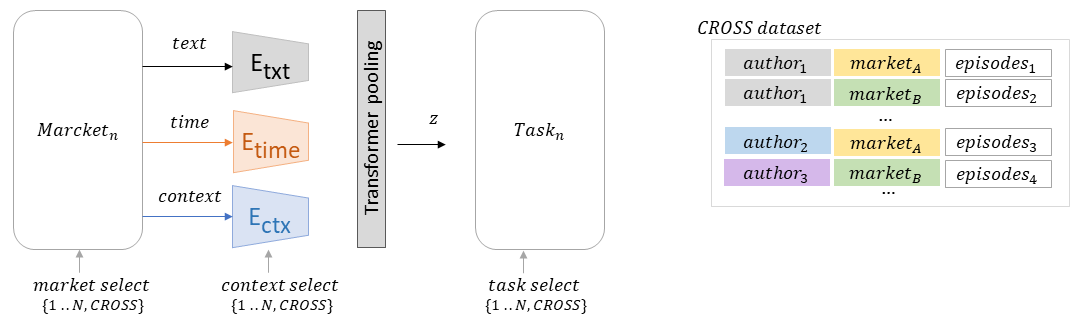}
    \caption{Multitask uses shared text and time embeddings $E_{txt}$ and $E_{time}$. Each market has its own context, $E_{ctx,n}; n \in \{1\dots N, \operatorname{CROSS}\}$. CROSS denotes the cross market dataset used to align inter-market embeddings.
    Model is based on SYSML\cite{maneriker2021sysml}}
    \label{fig_SYSMLCROSS}
\end{figure}

The singletask learning process, where we use a given single darkweb market $M_n$ for supervised author identification, using the model $f(\theta)$ under \acrshort{ALOSS} $L(\theta)$ by optimization, is straightforward. We term this combined authorship detection model implementation as \acrshort{AMODEL} to differentiate it from the proposed loss function \acrshort{ALOSS}. 

For multitask learning, where the aim is to detect users with different usernames across darkweb markets using style representations, we follow the approach proposed by SYSML\cite{maneriker2021sysml}, as shown in Figure~\ref{fig_SYSMLCROSS}. For each market $M_n$ we obtain the user representation $z$ using a shared text embedding $E_{txt}$ and time embedding $E_{time}$ model. However, each market $M_n$ has its own unique context embedding metric $E_{ctx,n}$, as the forum context is market dependent.
As shown on the right-side on Figure~\ref{fig_SYSMLCROSS}, to align the individual embeddings of the $E_{ctx,n} \forall n \in \{1, \dots, N\}$ we use a manually annotated subset of authors that were known to have migrated across markets. The authors common to two markets are provided a new label, e.g. $\operatorname{author}_1$, and the $\operatorname{CROSS}$ dataset and its associated $E_{ctx,CROSS}$ is used for embeddings.

\section{Experiments}

\subsection{darkweb datasets}
\begin{table*}[htbp]
    \centering
    \begin{tabular}{lrrrr} \toprule
         Market &  Train Posts & Test Posts & \#Users train & \#Users test \\ \midrule
         BMR & 30083 &30474 & 855 & 931\\
         Agora & 175978 & 179482 & 3115 & 4209\\ 
         SR2 & 373905 & 380779 &  5346 & 6580 \\
         SR & 379382 & 381959 & 6585 & 8865\\ \bottomrule
    \end{tabular}
    \caption{Statistics of the darkweb market discussion forum datasets.}
    \label{tab:dataset_stats}
\end{table*}

Our darkweb datasets are from \cite{munksgaard2016mixing}, who studied topic models on the forum posts across six large markets. We use \textit{Silk Road} (\textbf{SR}), \textit{Silk Road 2.0}~({\textbf{SR2}}), \textit{Agora Marketplace}~(\textbf{Agora}), and \textit{Black Market Reloaded} (\textbf{BMR}). 
Following \cite{maneriker2021sysml}, we ignore 'The Hub' as it is an 'omni-forum'~\cite{munksgaard2016mixing} discussing other marketplaces which is, as a result, significantly different layout and characteristics. Further, 'Evolution Marketplace' is excluded due to not having any PGP information to link users for the migration analysis. 

We use the pre-processing methodology from \cite{maneriker2021sysml}. Specifically, we use simple regex and rule based filters to replace quoted posts (i.e., posts that are begin replied to), PGP keys, PGP signatures, hashed messages, links, and images each with different special tokens (\texttt{[QUOTE]}, \texttt{[PGP PUBKEY]}, \texttt{[PGP SIGNATURE]}, \texttt{[PGP ENCMSG]}, \texttt{[LINK]}, \texttt{[IMAGE]}).
We use an episode length of 5 and consider users with sufficient posts for at least 2 episodes, \i.e., a threshold of users with posts $\geq$ 10. Information leak is prevented by splitting equal-sized train and test sets such that half the posts on the forum are before that time point. With this approach, test data can contain authors unseen during training.
Statistics of data after pre-processing is shown in Table~\ref{tab:dataset_stats}.

Based on prior research \cite{tai2019adversarial}, PGP keys are a strong indicator of shared authorship on darkweb markets. We use the manually verified and annotated authors based on PGP keys from \cite{maneriker2021sysml} which has 100 reliable labels, with 33 pairs matched as migrants across markets.

\subsection{Evaluation}
In our adopted dataset, no ground truth labels for a single author having multiple accounts is available.
Thus, we adopt an author identification posed as a retrieval based deep metric learning and euclidean search in the learned embedding space. It ranks all authors according to the relevance to an episode. 
Specifically, following \cite{maneriker2021sysml}, we compute the similarity of all episode embeddings $E = \{e_1, \dots e_n\}$ with a specific query embedding $q_i \subset E$, to obtain the similarity $R_i = \langle r_{i1}, r_{i2}, \dots r_{in} \rangle$. 

We use the following retrieval-based evaluation metrics: 

\textbf{Recall@k} (R@k): for an episode $e_i$, indicates if an episode by the same author is within the subset $ \langle r_{i1}, \dots, r_{ik} \rangle$ \cite{patel2022recall}. 
Specifically, R@k denotes the mean of these recall values over all the query samples.
\begin{equation}
    R@k = \frac{1}{\kappa}\sum\limits_{i=1}^{\kappa}{{1}_{\{\exists\ j | 1 \leq j \leq k, A(r_{ij}) = A(e_i)\}}}
\end{equation}

\textbf{Mean Reciprocal Rank} (MRR): evaluates a ranked list of answers to a query. 
Thus, the reciprocal rank for an episode is $\frac{1}{rank}$ of the first element (by similarity) with the same author. 
For multiple queries, $Q$, MRR is the mean of the reciprocal ranks. 
\begin{equation}
    MRR(Q) = \frac{1}{\kappa}\sum_{i=1}^\kappa \frac{1}{\min\limits_j  \left(A(r_{ij}) = A(e_i)\right)}
\end{equation}

We compare three baseline methods. 
First, we consider a CNN based character n-gram model for authorship attribution of short texts \cite{shrestha2017convolutional}, with modifications by \cite{maneriker2021sysml} to include time and context attributes (time, context). 
The second is IUR \cite{andrews2019learning}, which creates invariant user representations using softmax based metric learning and considers text, time and context. Our final benchmark is the current state-of-the-art, SYSML\cite{maneriker2021sysml}, which creates composite episodes of user activity from text, time and context for downstream softmax metric-based optimization. We evaluate all benchmarks for single dataset and multiple joint datasets (multitask), following \cite{maneriker2021sysml}.

\subsection{Results}
Results for episodes of length 5 are shown in Table~\ref{tab:baselines_comparison}. We report results for CNN, IUR and SYSML from \cite{maneriker2021sysml}. Our \acrshort{AMODEL} uses the proposed \acrshort{ALOSS} metric loss and uses the same deep neural network and hyperparameter settings for comparison. Moreover, our method is significantly easier to implement.

Following \cite{maneriker2021sysml}, we also provide results for comparison, namely w/o graph context and w/o time. While all models are better with these components, \acrshort{ALOSS} still shows less degradation over other models in each category of ablation.

For hyperpaprameters specific to our proposed loss, i.e. $\acrshort{ALOSS}(\operatorname{batch_size}, \tau, \alpha)$ we keep $\alpha=0.5$, and use the same batch\_size of 256 for the contrastive negatives as in \cite{maneriker2021sysml}. Additionally, as an ablation, we have included  \acrshort{AMODEL}(2048) which uses a $batch\_size=2048$, for convenience of the readers and to demonstrate how \acrshort{ALOSS} improves with batch size.

\begin{table*}
\centering
\begin{tabular}{lcccccccc} \toprule
	\multirow{2}{*}{Method}	&\multicolumn{2}{c}{BMR}	&	\multicolumn{2}{c}{Agora}	&	\multicolumn{2}{c}{SR2}	&	\multicolumn{2}{c}{SR}\\
			&MRR&	R@10&	MRR&	R@10&	MRR&	R@10&	MRR&	R@10\\ \midrule
	\cite{shrestha2017convolutional} (CNN) &	0.070	&	0.165	&	0.126	&	0.214	&	0.082	&	0.131	&	0.036	&	0.073	\\
	 + time + context &	0.235	&	0.413	&	0.152	&	0.263	&	0.118	&	0.210	&	0.094	&	0.178	\\
	 + time + context + transformer pooling &	0.219	&	0.409	&	0.146	&	0.266	&	0.117	&	0.207	&	0.113	&	0.205	\\ 
\hline
	\cite{andrews2019learning} (IUR) \\
	 using mean pooling &	0.223	&	0.408	&	0.114	&	0.218	&	0.126	&	0.223	&	0.109	&	0.190	\\
	 using transformer pooling &	0.283	&	0.477	&	0.127	&	0.234	&	\rm{0.130}	&	\rm{0.229}	&	0.118	&	0.204	\\
\hline
	\cite{maneriker2021sysml} SYSML - singletask &	\rm{0.320}	&	\rm{0.533}	&	\rm{0.152}	&	\rm{0.279}	&	0.123	&	0.210	&	\rm{0.157}	&	\rm{0.266}	\\
	 - graph context &	0.265	&	0.454	&	0.144	&	0.251	&	0.089	&	0.150	&	0.049	&	0.094	\\
	 - graph context - time &	0.277	&	0.477	&	0.123	&	0.198	&	0.079	&	0.131	&	0.04	&	0.080	\\
\hline
	\cite{maneriker2021sysml} SYSML - multitask &	\rm{0.438}	&	\rm{0.642}	&	\rm{0.303}	&	\rm{0.466}	&	\rm{0.304}	&	\rm{0.464}	&	\rm{0.227}	&	\rm{0.363}	\\
	 - graph context &	0.396	&	0.602	&	\rm{0.308}	&	\rm{0.469}	&	0.293	&	0.442	&	0.214	&	0.347	\\
	 - graph context - time &	0.366	&	0.575	&	0.251	&	0.364	&	0.236	&	0.358	&	0.167	&	0.280	\\
\bottomrule \bottomrule
    
    Ours - \acrshort{AMODEL} - singletask & \cellcolor[HTML]{E1C9F6}{0.3275}  &  \cellcolor[HTML]{E1C9F6}{0.514}   &   \cellcolor[HTML]{E1C9F6}{0.2796}  &  \cellcolor[HTML]{E1C9F6}{0.439}    &   \cellcolor[HTML]{E1C9F6}{0.2574}  &  \cellcolor[HTML]{E1C9F6}{0.409}  &     \cellcolor[HTML]{E1C9F6}{0.2075}  &  \cellcolor[HTML]{E1C9F6}{0.335} \\
    - graph context               & 0.2520   & 0.435  &    0.2379  &  0.406   &    0.2137  &  0.349    &   0.1678  &  0.281 \\
    - graph context - time        &  0.2878  &  0.491  &    0.2334  &  0.349   &    0.2131  &  0.328    &   0.1718  &  0.293 \\
\bottomrule
    
    Ours - \acrshort{AMODEL} - multitask & \textbf{0.4846}  &  \textbf{0.683}   &      \textbf{0.3491}  &  \textbf{0.505}   &   \textbf{0.3400}  &   \textbf{0.491}  &   \textbf{0.2644}  &  \textbf{0.416} \\
    - graph context               & 0.4354   &  0.618    &   0.3259  &  0.470      & 0.3143  &   0.473  &   0.2361  &  0.371 \\
    - graph context - time        & 0.4042 &    0.607   &    0.2936  &  0.406      & 0.2931  &   0.403   &  0.2019 &   0.329 \\
    
    Ours - \acrshort{AMODEL} - multitask(2048) & \textbf{0.6512}  &  \textbf{0.768}   &      \textbf{0.3886}  &  \textbf{0.539}   &   \textbf{0.3472}  &   \textbf{0.500}  &   \textbf{0.2935}  &  \textbf{0.448} \\
\bottomrule
\end{tabular}
\caption{Best performing multitask results in \textbf{bold}.
All models use $batch\_size=256$, with the exception of \acrshort{AMODEL}(2048) which uses a $batch\_size=2048$. 
Best performing singletask results are \colorbox[HTML]{E1C9F6}{highlighted}. All  $\sigma_{MRR}< 0.02$, $\sigma_{R@10} < 0.03$.
We use $\tau = 0.2$ for singletask and $\tau = 0.3$ for the muti-task.
In all metrics, higher is better. Our novel combined loss function improves both singletask and multitask performance over state-of-the-art \cite{maneriker2021sysml}. Additionally, the novel loss on the multitask cross-market setup also shows significant consistent improvements in over singletask performance.}
\label{tab:baselines_comparison}
\end{table*}

\subsection{Effect of hyperparameters $\tau$ and batch\_size}
In the following sections we look in to \acrshort{ALOSS} Hyperparameter Stability. 

\begin{figure}[p]
    \centering \includegraphics[width=1.0\columnwidth]{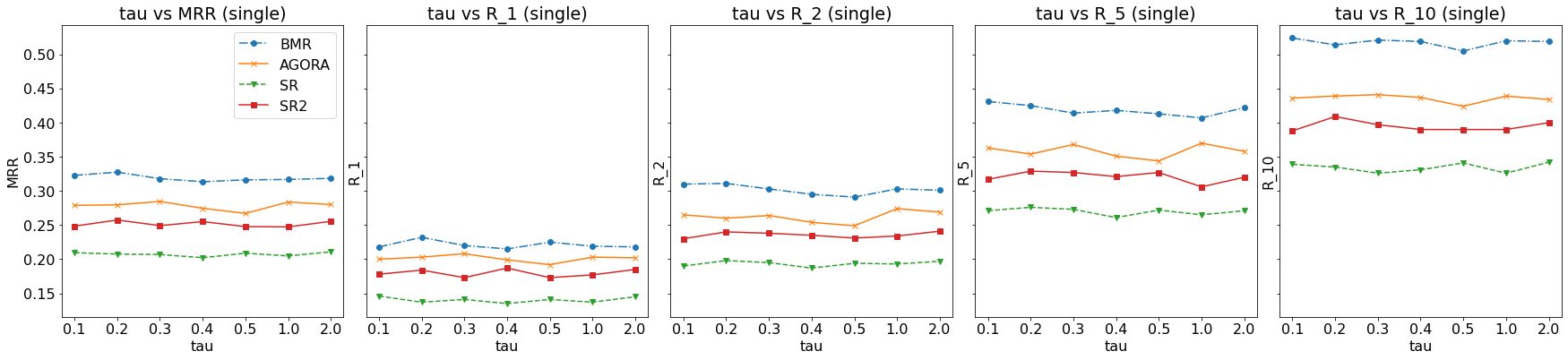}
    \caption{$\tau$ temperature effect on singletask performance indicates that the model is robust to a wide range of values. Hence there is no requirement to use an optimal value.}
    \label{fig_singletau}
\end{figure}
\begin{figure}
    \centering \includegraphics[width=1.0\columnwidth]{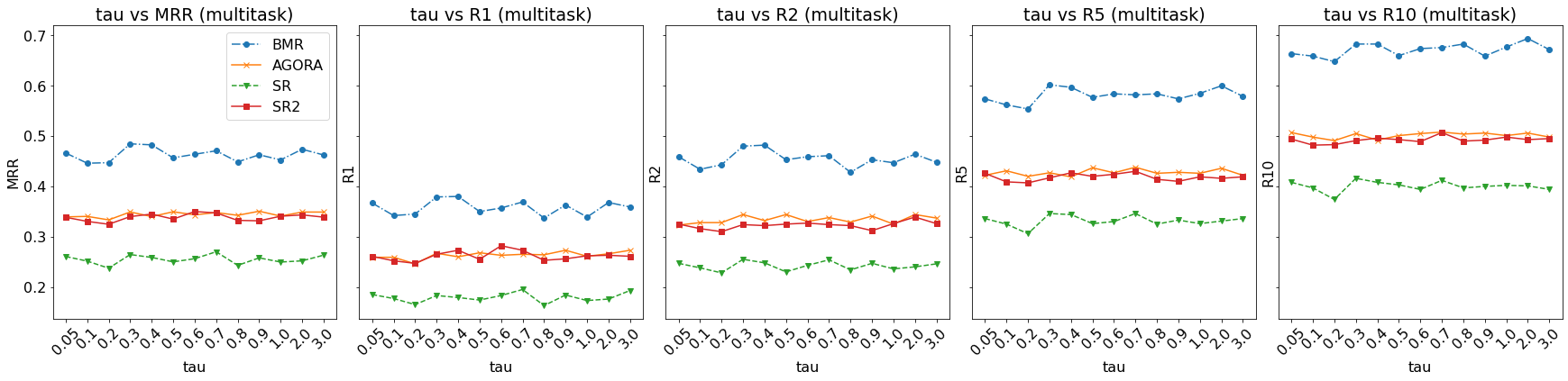}
    \caption{$\tau$ temperature effect on multitask performance indicates that the model is robust to a wide range of values. Hence there is no requirement to use an optimal value.}
    \label{fig_multitau}
\end{figure}
We found that $\tau=0.2$ to be marginally better and $\tau=0.3$ to be marginally better for single and multitask models respectively. However, as seen in Figure~\ref{fig_singletau} for singletask and Figure~\ref{fig_multitau}, $\tau$ has less significant effect on performance indicates that the model is robust to a wide range of values. Hence there is no requirement to use an optimal value.
Our findings are similar to \cite{khosla2020supervised}, which reported $\tau=0.1$ as optimal. Smaller temperature ($\tau$) values are better than larger values, while extremely low temperatures are harder to train due to numerical instability.

\begin{figure}
    \centering \includegraphics[width=1.0\columnwidth]{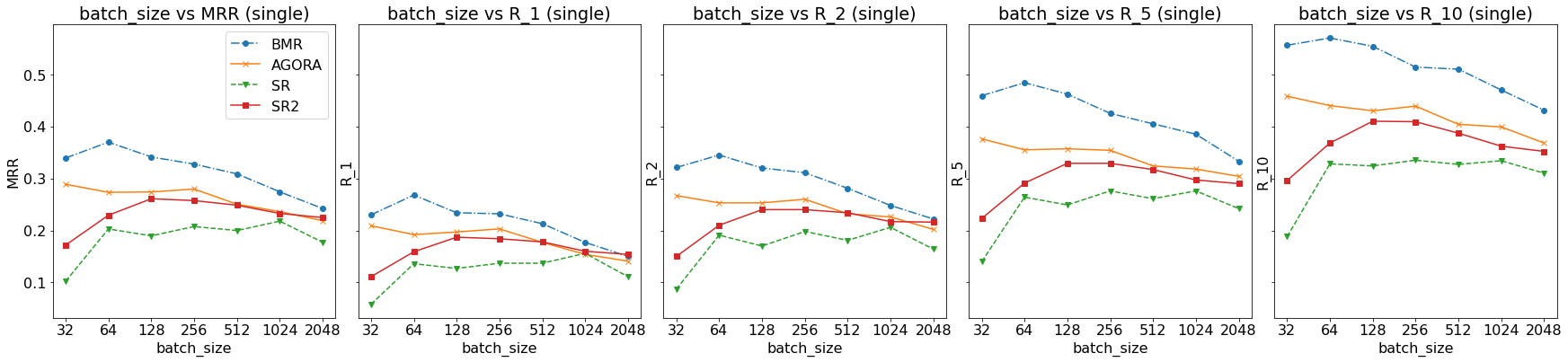}
    \caption{singletask performance with varying $batch\_size$}
    \label{fig_singleBatch}
\end{figure}
\begin{figure}
    \centering \includegraphics[width=1.0\columnwidth]{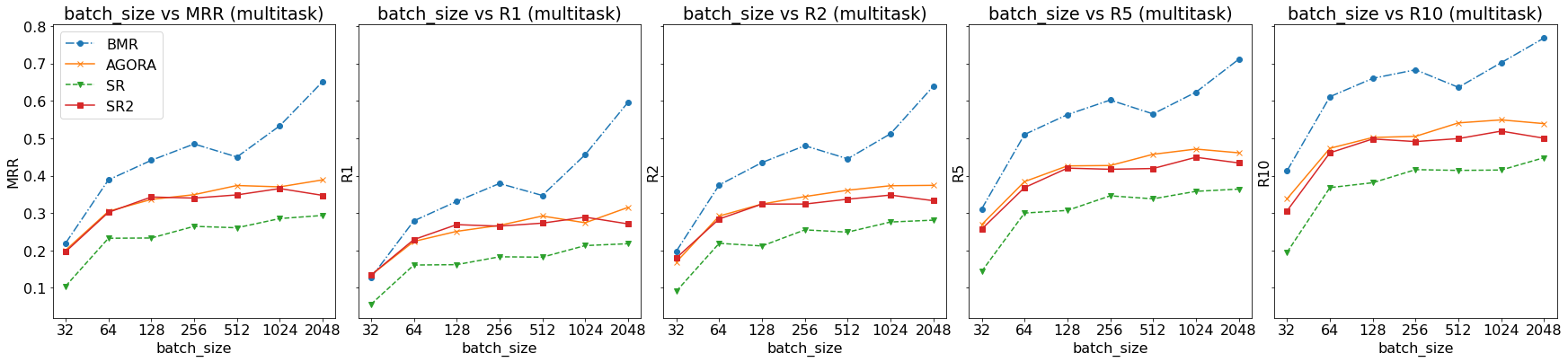}
    \caption{In multitask there is a consistent increase in performance as the $batch\_size$ increases.}
    \label{fig_multiBatch}
\end{figure}
Our results show different results for single and multitask models in relation to batch\_size. Figure~\ref{fig_singleBatch} shows that a larger batch\_size results in slightly lower singletask performance in BMR and AGORA while in SR and SR2 a peek is achieved around 64 and 128, though overall the performance is increasing. 

In multitask models, as shown in Figure~\ref{fig_multiBatch}, there is a consistent increase in performance as the batch\_size increases. Specifically, we achieve 0.6512 MRR on BMR, a 49\% increase, 0.3889 MRR on AGORA, a 28\% increase, 0.2935 MRR on SR, a 23\% increase and 0.3471 MRR on SR2 a 14\% increase over current bester performing model, SYSML. We attribute the increase in performance due to the better discriminative ability achieved by our \acrshort{ALOSS}. 

This agrees with the theoretical work in \cite{wang2020understanding} and the numerical results in \cite{he2020momentum} (e.g., batch size of 65536), where large values of $batch\_size$ lead to higher performance in downstream tasks. For supervised (non-block based, traditional) contrastive loss \cite{khosla2020supervised} used batch sizes of 6144, while a batch size of 2048 was also sufficient. Thus \acrshort{ALOSS} does not require adversely large batch sizes, and works well with micro batches while improving significantly with larger batches, especially for multitask learning.

\subsection{Comparison with other softmax improvement approaches}
\begin{figure}
    \centering \includegraphics[width=0.5\columnwidth]{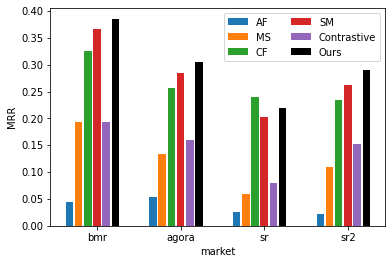}
    \caption{Metric loss performance comparison. \acrshort{ALOSS}, with negative block contrasting performs well in 3 of the 4 cases, while contrastive loss gives poor performance alone. We reproduce the results for AF,MS,CF and SM from \cite{maneriker2021sysml}. Results show average performance (MRR) over $batch\_size = \{1,3,5\}$  }
    \label{fig_NCE}
\end{figure}
Figure~\ref{fig_NCE} shows various recently proposed metric learning methods compared with our \acrshort{ALOSS}. As can be seen, supervised (traditional) pairwise contrastive loss performs poorly in its unmodified form. However, our \acrshort{ALOSS} using negative block contrastive loss is able to achieve best results. \acrshort{ALOSS} is also better than the traditional softmax (SM) in very dataset.

\subsection{Effect of episode size}
\begin{figure}
    \centering \includegraphics[width=1.0\columnwidth]{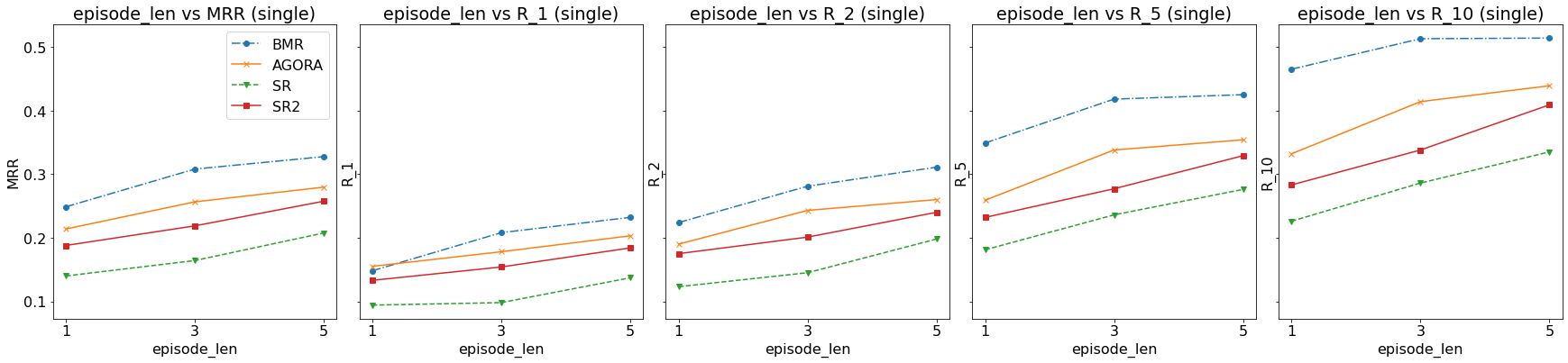}
    \caption{Larger episode sizes improve singletask performance across all datasets.}
    \label{fig_singleepisodelength}
\end{figure}
\begin{figure}
    \centering \includegraphics[width=1.0\columnwidth]{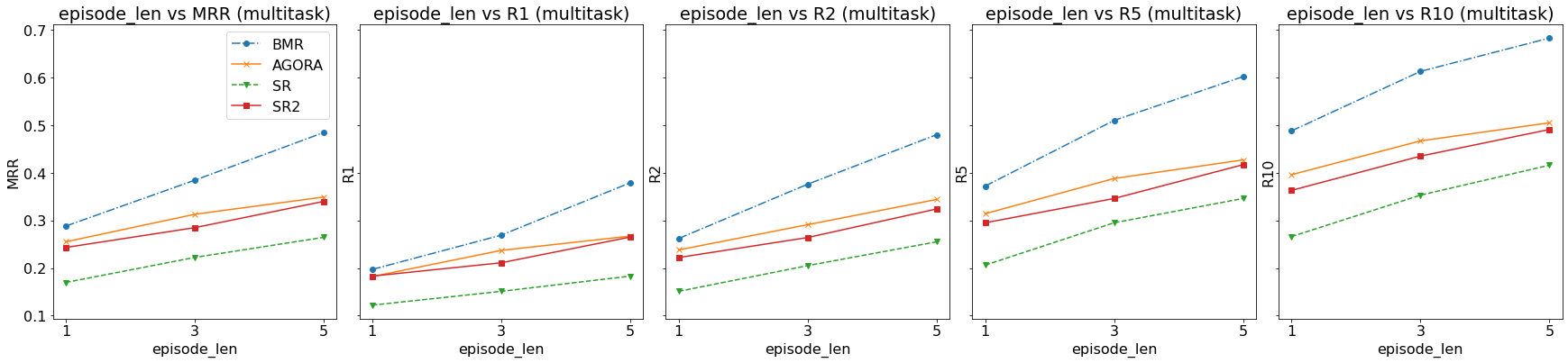}
    \caption{Larger episode sizes improve multitask performance across all datasets.}
    \label{fig_multiepisodelength}
\end{figure}
Figure~\ref{fig_singleepisodelength} and Figure~\ref{fig_multiepisodelength} shows increases with episode lengths, as the model is able to use more linguistic information with larger amounts of episodes.  Increasing the episode lengths while keeping all other hyperparameters constant results in consistently better performance with \acrshort{ALOSS}

\begin{figure}
    \centering \includegraphics[width=0.5\columnwidth]{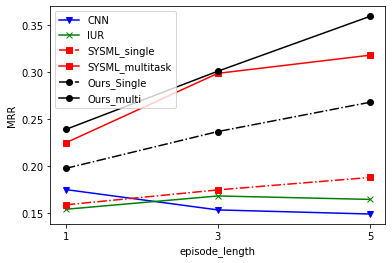}
    \caption{Under all episode lengths \{1,3,5\}, compared to the baselines, \acrshort{ALOSS} is able to achieve better performance, consistently outperforming traditional softmax based models (SYSML).}
    \label{fig_EPISODELEN}
\end{figure}
In order to benchmark across other models, we also compare different models under varying episode lengths in Figure~\ref{fig_EPISODELEN}. Under all episode sizes, compared to the baselines, \acrshort{ALOSS} is able to achieve better performance, and shows increasing gains over the next best model, SYSML. As the model difference is mainly in our use of \acrshort{ALOSS}, we attribute the improvements to its ability to learn better representations.

\section{Conclusion}
In this paper, we introduced a novel \acrshort{ALOSS} that uses negative block contrastive loss to improve the discriminative power of softmax loss for classification. Different from prior margin-based approaches, we address the problem via uniformly distributing points on a hyperspace by minimizing the total \textit{sub-batched block wise}, or prototype, potential w.r.t. a certain kernel function. 
Our \acrshort{ALOSS} is augmentation-free, does not require sampling, is robust to hyperparameter variations, scalable and relatively easier to implement or retro fit.
Experimentally, we demonstrate the application of the \acrshort{ALOSS} on 4 real darkweb datasets in author and sybil detection tasks and achieve state-ot-the-art. 

In future work, we will focus on exploring the effect of $batch\_size$ on singletask performance. Additionally, motivated by the success of hard-negative sampling strategies, we will extend the work to use hard negatives as a means to improve computational efficiency and classification accuracy. As a natural extension, future work can explore the use of \acrshort{ALOSS} on other domains, e.g. face detection in vision, and designing better $f(\theta)$ model architectures. 
Finally, a key question is the effectiveness of authorship attribution in the presence of adversarial content, where users are actively attempting to mask or obfuscate intentions and stylographic cues, in addition to sybil accounts.

\section*{Acknowledgments}
Dedicated to Sugandi.

\clearpage
\bibliographystyle{unsrt}  
\bibliography{main} 

\begin{thebibliography}{10}

\bibitem{kumar2020edarkfind}
Ramnath Kumar, Shweta Yadav, Raminta Daniulaityte, Francois Lamy, Krishnaprasad
  Thirunarayan, Usha Lokala, and Amit Sheth.
\newblock edarkfind: Unsupervised multi-view learning for sybil account
  detection.
\newblock In {\em Proceedings of The Web Conference 2020}, pages 1955--1965,
  2020.

\bibitem{manolache2022veridark}
Andrei Manolache, Florin Brad, Antonio Barbalau, Radu~Tudor Ionescu, and Marius
  Popescu.
\newblock Veridark: A large-scale benchmark for authorship verification on the
  dark web.
\newblock In {\em Thirty-sixth Conference on Neural Information Processing
  Systems Datasets and Benchmarks Track}, 2022.

\bibitem{munksgaard2016mixing}
Rasmus Munksgaard and Jakob Demant.
\newblock Mixing politics and crime--the prevalence and decline of political
  discourse on the cryptomarket.
\newblock {\em International Journal of Drug Policy}, 35:77--83, 2016.

\bibitem{biryukov2014content}
Alex Biryukov, Ivan Pustogarov, Fabrice Thill, and Ralf-Philipp Weinmann.
\newblock Content and popularity analysis of tor hidden services.
\newblock In {\em 2014 IEEE 34th International Conference on Distributed
  Computing Systems Workshops (ICDCSW)}, pages 188--193. IEEE, 2014.

\bibitem{elbahrawy2020collective}
Abeer ElBahrawy, Laura Alessandretti, Leonid Rusnac, Daniel Goldsmith,
  Alexander Teytelboym, and Andrea Baronchelli.
\newblock Collective dynamics of dark web marketplaces.
\newblock {\em Scientific reports}, 10(1):1--8, 2020.

\bibitem{ebrahimi2018detecting}
Mohammadreza Ebrahimi, Mihai Surdeanu, Sagar Samtani, and Hsinchun Chen.
\newblock Detecting cyber threats in non-english dark net markets: A
  cross-lingual transfer learning approach.
\newblock In {\em 2018 IEEE International Conference on Intelligence and
  Security Informatics (ISI)}, pages 85--90. IEEE, 2018.

\bibitem{lorenzo2018know}
Nuria Lorenzo-Dus and Matteo Di~Cristofaro.
\newblock ‘i know this whole market is based on the trust you put in me and i
  don’t take that lightly’: Trust, community and discourse in crypto-drug
  markets.
\newblock {\em Discourse \& Communication}, 12(6):608--626, 2018.

\bibitem{maneriker2021sysml}
Pranav Maneriker, Yuntian He, and Srinivasan Parthasarathy.
\newblock Sysml: Stylometry with structure and multitask learning: Implications
  for darknet forum migrant analysis.
\newblock In {\em Proceedings of the 2021 Conference on Empirical Methods in
  Natural Language Processing}, pages 6844--6857, 2021.

\bibitem{tai2019adversarial}
Xiao~Hui Tai, Kyle Soska, and Nicolas Christin.
\newblock Adversarial matching of dark net market vendor accounts.
\newblock In {\em Proceedings of the 25th ACM SIGKDD International Conference
  on Knowledge Discovery \& Data Mining}, pages 1871--1880, 2019.

\bibitem{acnikevivcs2021relationship}
Germans A{\c{n}}ikevi{\v{c}}s.
\newblock Relationship between vendor popularity and prices on dark web
  marketplaces.
\newblock {B.S.} thesis, University of Twente, 2021.

\bibitem{shrestha2017convolutional}
Prasha Shrestha, Sebastian Sierra, Fabio~A Gonzelez, Paolo Rosso, Manuel
  Montes-y Gomez, and Thamar Solorio.
\newblock Convolutional neural networks for authorship attribution of short
  texts.
\newblock {\em EACL 2017}, page 669, 2017.

\bibitem{andrews2019learning}
Nicholas Andrews and Marcus Bishop.
\newblock Learning invariant representations of social media users.
\newblock In {\em Proceedings of the 2019 Conference on Empirical Methods in
  Natural Language Processing and the 9th International Joint Conference on
  Natural Language Processing (EMNLP-IJCNLP)}, pages 1684--1695, 2019.

\bibitem{wang2018additive}
Feng Wang, Jian Cheng, Weiyang Liu, and Haijun Liu.
\newblock Additive margin softmax for face verification.
\newblock {\em IEEE Signal Processing Letters}, 25(7):926--930, 2018.

\bibitem{wen2016discriminative}
Yandong Wen, Kaipeng Zhang, Zhifeng Li, and Yu~Qiao.
\newblock A discriminative feature learning approach for deep face recognition.
\newblock In {\em European conference on computer vision}, pages 499--515.
  Springer, 2016.

\bibitem{wang2018cosface}
Hao Wang, Yitong Wang, Zheng Zhou, Xing Ji, Dihong Gong, Jingchao Zhou, Zhifeng
  Li, and Wei Liu.
\newblock Cosface: Large margin cosine loss for deep face recognition.
\newblock In {\em Proceedings of the IEEE conference on computer vision and
  pattern recognition}, pages 5265--5274, 2018.

\bibitem{deng2019arcface}
Jiankang Deng, Jia Guo, Niannan Xue, and Stefanos Zafeiriou.
\newblock Arcface: Additive angular margin loss for deep face recognition.
\newblock In {\em Proceedings of the IEEE/CVF conference on computer vision and
  pattern recognition}, pages 4690--4699, 2019.

\bibitem{wang2019multi}
Xun Wang, Xintong Han, Weilin Huang, Dengke Dong, and Matthew~R Scott.
\newblock Multi-similarity loss with general pair weighting for deep metric
  learning.
\newblock In {\em Proceedings of the IEEE/CVF Conference on Computer Vision and
  Pattern Recognition}, pages 5022--5030, 2019.

\bibitem{kulatilleke2022efficient}
Gayan~K Kulatilleke, Marius Portmann, and Shekhar~S Chandra.
\newblock Efficient block contrastive learning via parameter-free meta-node
  approximation.
\newblock {\em arXiv preprint arXiv:2209.14067}, 2022.

\bibitem{kulatilleke2022scgc}
Gayan~K Kulatilleke, Marius Portmann, and Shekhar~S Chandra.
\newblock Scgc: Self-supervised contrastive graph clustering.
\newblock {\em arXiv preprint arXiv:2204.12656}, 2022.

\bibitem{wang2020understanding}
Tongzhou Wang and Phillip Isola.
\newblock Understanding contrastive representation learning through alignment
  and uniformity on the hypersphere.
\newblock In {\em International Conference on Machine Learning}, pages
  9929--9939. PMLR, 2020.

\bibitem{logeswaran2018efficient}
Lajanugen Logeswaran and Honglak Lee.
\newblock An efficient framework for learning sentence representations.
\newblock In {\em International Conference on Learning Representations}, 2018.

\bibitem{gunel2020supervised}
Beliz Gunel, Jingfei Du, Alexis Conneau, and Veselin Stoyanov.
\newblock Supervised contrastive learning for pre-trained language model
  fine-tuning.
\newblock In {\em International Conference on Learning Representations}, 2020.

\bibitem{patel2022recall}
Yash Patel, Giorgos Tolias, and Ji{\v{r}}{\'\i} Matas.
\newblock Recall@ k surrogate loss with large batches and similarity mixup.
\newblock In {\em Proceedings of the IEEE/CVF Conference on Computer Vision and
  Pattern Recognition}, pages 7502--7511, 2022.

\bibitem{liu2017sphereface}
Weiyang Liu, Yandong Wen, Zhiding Yu, Ming Li, Bhiksha Raj, and Le~Song.
\newblock Sphereface: Deep hypersphere embedding for face recognition.
\newblock In {\em Proceedings of the IEEE conference on computer vision and
  pattern recognition}, pages 212--220, 2017.

\bibitem{chen2020simple}
Ming Chen, Zhewei Wei, Zengfeng Huang, Bolin Ding, and Yaliang Li.
\newblock Simple and deep graph convolutional networks.
\newblock In {\em International Conference on Machine Learning}, pages
  1725--1735. PMLR, 2020.

\bibitem{khosla2020supervised}
Prannay Khosla, Piotr Teterwak, Chen Wang, Aaron Sarna, Yonglong Tian, Phillip
  Isola, Aaron Maschinot, Ce~Liu, and Dilip Krishnan.
\newblock Supervised contrastive learning.
\newblock {\em Advances in Neural Information Processing Systems},
  33:18661--18673, 2020.

\bibitem{arora2019theoretical}
Sanjeev Arora, Hrishikesh Khandeparkar, Mikhail Khodak, Orestis Plevrakis, and
  Nikunj Saunshi.
\newblock A theoretical analysis of contrastive unsupervised representation
  learning.
\newblock In {\em 36th International Conference on Machine Learning, ICML
  2019}, pages 9904--9923. International Machine Learning Society (IMLS), 2019.

\bibitem{kim2014convolutional}
Yoon Kim.
\newblock Convolutional neural networks for sentence classification.
\newblock In {\em EMNLP}, 2014.

\bibitem{suarez2018tutorial}
Juan~Luis Su{\'a}rez-D{\'\i}az, Salvador Garc{\'\i}a, and Francisco Herrera.
\newblock A tutorial on distance metric learning: Mathematical foundations,
  algorithms, experimental analysis, prospects and challenges (with appendices
  on mathematical background and detailed algorithms explanation).
\newblock {\em arXiv preprint arXiv:1812.05944}, 2018.

\bibitem{dong2017metapath2vec}
Yuxiao Dong, Nitesh~V Chawla, and Ananthram Swami.
\newblock metapath2vec: Scalable representation learning for heterogeneous
  networks.
\newblock In {\em Proceedings of the 23rd ACM SIGKDD international conference
  on knowledge discovery and data mining}, pages 135--144, 2017.

\bibitem{devlin-etal-2019-bert}
Jacob Devlin, Ming-Wei Chang, Kenton Lee, and Kristina Toutanova.
\newblock {BERT}: Pre-training of deep bidirectional transformers for language
  understanding.
\newblock In {\em Proceedings of the 2019 Conference of the North {A}merican
  Chapter of the Association for Computational Linguistics: Human Language
  Technologies, Volume 1 (Long and Short Papers)}, pages 4171--4186,
  Minneapolis, Minnesota, June 2019. Association for Computational Linguistics.

\bibitem{vaswani2017attention}
Ashish Vaswani, Noam Shazeer, Niki Parmar, Jakob Uszkoreit, Llion Jones,
  Aidan~N Gomez, {\L}ukasz Kaiser, and Illia Polosukhin.
\newblock Attention is all you need.
\newblock {\em Advances in neural information processing systems}, 30, 2017.

\bibitem{kulatilleke2021fdgatii}
Gayan~K Kulatilleke, Marius Portmann, Ryan Ko, and Shekhar~S Chandra.
\newblock Fdgatii: Fast dynamic graph attention with initial residual and
  identity mapping.
\newblock {\em arXiv preprint arXiv:2110.11464}, 2021.

\bibitem{he2020momentum}
Kaiming He, Haoqi Fan, Yuxin Wu, Saining Xie, and Ross Girshick.
\newblock Momentum contrast for unsupervised visual representation learning.
\newblock In {\em Proceedings of the IEEE/CVF conference on computer vision and
  pattern recognition}, pages 9729--9738, 2020.

\end{thebibliography}
\end{document}